\documentclass[conference]{IEEEtran}

\usepackage{times}
\usepackage{epsfig}
\usepackage{graphicx}
\usepackage{caption}
\usepackage{subcaption}

\ifCLASSINFOpdf
\else
\fi
\usepackage[cmex10]{amsmath}
\usepackage{algorithm}
\usepackage[noend]{algpseudocode}

\algrenewcommand\algorithmicforall{\textbf{foreach}}
\algrenewcommand\algorithmicindent{.8em}

\algrenewcommand{\algorithmicrequire}{\textbf{Input:}}
\algrenewcommand{\algorithmicensure}{\textbf{Output:}}
\begin{document}
%
% paper title
% Titles are generally capitalized except for words such as a, an, and, as,
% at, but, by, for, in, nor, of, on, or, the, to and up, which are usually
% not capitalized unless they are the first or last word of the title.
% Linebreaks \\ can be used within to get better formatting as desired.
% Do not put math or special symbols in the title.

%\title{Trace Lasso Regularized L1-norm Graph Cut}% for Classification of Hyperspectral Images}
\title{A Trace Lasso Regularized L1-norm Graph Cut for Highly Correlated Noisy Hyperspectral Image}
%
%
% author names and IEEE memberships
% note positions of commas and nonbreaking spaces ( ~ ) LaTeX will not break
% a structure at a ~ so this keeps an author's name from being broken across
% two lines.
% use \thanks{} to gain access to the first footnote area
% a separate \thanks must be used for each paragraph as LaTeX2e's \thanks
% was not built to handle multiple paragraphs
\author{\IEEEauthorblockN{Ramanarayan Mohanty}
\IEEEauthorblockA{ATDC, 
IIT Kharagpur,\\ %India 721302\\
Email: ramanarayan@iitkgp.ac.in}
\and
\IEEEauthorblockN{S L Happy}
\IEEEauthorblockA{Dept. of EE, 
IIT Kharagpur, \\%India 721302\\
Email: happy@iitkgp.ac.in }
\and
\IEEEauthorblockN{Nilesh Suthar}
\IEEEauthorblockA{{Dept. of Biotechnology}\\ 
IIT Kharagpur,} %India 721302}\\
%Email: nsnilesh071@gmail.com}
\and
\IEEEauthorblockN{Aurobinda Routray}
\IEEEauthorblockA{Dept. of EE, 
IIT Kharagpur, \\ %India 721302\\
Email: aroutray@iitkgp.ac.in }}

\maketitle

% As a general rule, do not put math, special symbols or citations
% in the abstract or keywords.
\begin{abstract}
This work proposes an adaptive trace lasso regularized L1-norm based graph cut method for dimensionality reduction of Hyperspectral images, called as `Trace Lasso-L1 Graph Cut' (TL-L1GC). The underlying idea of this method is to generate the optimal projection matrix by considering both the sparsity as well as the correlation of the data samples. The conventional L2-norm used in the objective function is sensitive to noise and outliers. Therefore, in this work L1-norm is utilized as a robust alternative to L2-norm. Besides, for further improvement of the results, we use a penalty function of trace lasso with the L1GC method. It adaptively balances the L2-norm and L1-norm simultaneously by considering the data correlation along with the sparsity. We obtain the optimal projection matrix by maximizing the ratio of between-class dispersion to within-class dispersion using L1-norm with trace lasso as the penalty. Furthermore, an iterative procedure for this TL-L1GC method is proposed to solve the optimization function. The effectiveness of this proposed method is evaluated on two benchmark HSI datasets.
\end{abstract}

% Note that keywords are not normally used for peerreview papers.

% For peer review papers, you can put extra information on the cover
% page as needed:
% \ifCLASSOPTIONpeerreview
% \begin{center} \bfseries EDICS Category: 3-BBND \end{center}
% \fi
%
% For peerreview papers, this IEEEtran command inserts a page break and
% creates the second title. It will be ignored for other modes.
\IEEEpeerreviewmaketitle

\section{Introduction}
Hyperspectral remote sensing images (HSI) capture the inherent properties of the observed scene through hundreds of contiguous spectral bands. These spectral bands are highly correlated \cite{zabalza2015novel} and rich with information contents. The captured information of the spectral bands offer great potentials for information retrieval. However, these large number of correlated spectral bands contain redundant information. This high dimensional redundant data is a significant challenge in HSI information extraction for the subsequent classification task. Spectral domain dimensionality reduction (DR) method is widely employed in HSI data analysis to mitigate this challenge.
%To mitigate this challenge, spectral domain dimensionality reduction (DR) method is widely employed in HSI data analysis. 
%An effective DR method seeks a low dimensional representation of high dimension data with most vital information content \cite{zhang2013semisupervised}. 
It improves the classification performance by reducing computational complexity as well as storage capacity and explores the intrinsic property of the extracted features.     

%Hyperspectral remote sensing images with high spatial and spectral resolution is used to capture the inherent properties of the surface. These hyperspectral images (HSI) contains a huge number of contiguous spectral bands. It spreads over a narrow spectral bandwidth with wealth of information content. These informations are used for characterization, identification and classification of the physical and chemical properties of the land-cover with improved accuracy. The huge spectral bands implies the high dimensional redundant HSI data. This high dimensionality is the major challenge in HSI classification. In order to overcome this challenge, dimensionality reduction (DR) is usually applied to the HSI data. An effective DR method reduces the high dimension data into low dimensional representative features. This DR method improves the classification performance by reducing computational complexity and exploring the intrinsic property of the reduced data features.

For HSI analysis, a wide variety of DR methods have drawn attention. There are broadly two types of DR methods, unsupervised and supervised. Among them, principal component analysis (PCA) \cite{martinez2001pca} is widely used in the unsupervised category. Whereas, in supervised category linear discriminant analysis (LDA) \cite{ye2004optimization} is the most popular one, which uses the labeled samples to determine the projection matrix. However, LDA considers the class center for its computation by assuming the data distribution as Gaussian. Hence, it fails to handle the real world HSI data with more complex distribution. The graph-based scaling cut (SC) method proposed in \cite{zhang2009local}, \cite{zhang2015scaling} address the issue mentioned earlier. This graph-based method addresses the issue by constructing a pairwise similarity matrix among the samples of the classes.
%The aforementioned issue was addressed by a graph based scaling cut (SC) method proposed in \cite{zhang2009local}, \cite{zhang2015scaling}. This graph based method addressed this issue by constructing a pairwise similarity matrix among the samples of the classes.

The idea behind the graph based SC method is to project the data into a lower dimensional space by maximizing the between-class variance with respect to the total-class variance. However, the SC method by Zhang \textit{et al.} in \cite{zhang2015scaling} are sensitive in handling the outliers and noise in the dataset, because the SC method uses the conventional L2-norm to compute the dissimilarity matrix between the samples. The square operation in the L2-norm exaggerates the effects of the outliers and noises in the data. 

Generally, L1-norm based DR method is a robust alternative to the L2-norm based method to handle the outliers problem \cite{wang2014fisher, liu2017non, mohanty2017graph, ke2005robust, kwak2008principal, li2010l1}. Li \textit{et.al.} in \cite{li2015robust} proposed the 2D version of the LDA (L1-2DLDA) using the L1-norm optimization. Similarly, in \cite{wang2014fisher}, Wang \textit{et.al.} proposed LDA-L1 by solving the supervised LDA method  iteratively using L1-norm maximization. In \cite{mohanty2017graph}, Mohanty \textit{et.al.} computed the dissimilarity matrix using L1-norm and proposed the L1-SC method by greedy strategy. However, the L1-norm have an undesired stability problem when the data are strongly correlated with each other \cite{grave2011trace}. For example, if two variables are correlated, the L1-norm based method will only choose one of them randomly without considering any correlation metric.

The L1-norm based SC method in \cite{mohanty2017graph} overlooks the spectral correlation property of HSI data, which adversely affect its performance. The addition of a convex penalty or regularization term (squared L2-norm) \cite{zou2005regularization} to the L1-norm based objective function is the best possible remedy for the instability issue \cite{grave2011trace}. However, these squared L2-norm based regularization terms are blind to the exact correlation structure of the data. That means this squared regularization term is not uniformly required for every variable or the regularizer is not adaptive. Hence, ``trace lasso" was proposed in \cite{grave2011trace} as an adaptive regularizer to keep the balance between L1 and L2-norm by considering both the sparsity and correlation of the data.

%, while   i.e. the regularizer is adaptive.   To address this instability issue, penalty is          
% The conventional SC works by computing the dissimilarity matrix among the data samples. This dissimilarity matrix computation is mostly done by calculating the conventional L2-norm between the samples. The square operation in L2-norm criterion magnifies the outliers \cite{ding2006r}, \cite{wang2014fisher}. Therefore, the presence of outliers drift the projection vectors from the desired projection direction. Hence, dimension reduction and classification of hyperspectral data demands robust algorithms that are resistant to possible outliers. 

%It is found that, the L1-norm based DR method is a robust alternative to handle outliers problem in image classification  computed the covariance matrix using L1-norm and proposed PCA-L1 by greedy strategy. Ke and Kanade, in \cite{ke2005robust}, proposed L1-PCA by using the alternative convex method to solve the projection matrix.  Similarly 

%SC has proved its worth in HSI classification \cite{zhang2015scaling}. 
L1-norm based SC has achieved excellent performance for noisy HSI data classification in \cite{mohanty2017graph}. The trace lasso (TL) regularizer has proved its worth on feature extraction of image datasets in \cite{lu2016l1}. However, our HSI data is noisy and has the strong spectral correlation. Furthermore, the high-resolution HSI data exhibit high spatial variability. That lead to poor spatial pixel correlation resulting undesired classification results. Hence, pixel consistency is essential for maintaining spatial correlation in HSI data. Motivated by the existing literature, we propose a trace lasso regularized L1-norm based graph cut method (TL-L1GC) preceded by a guided filter \cite{kang2014spectral} for DR and classification of noisy and correlated HSI data. In this work, we formulate the GC algorithm into an L1-norm optimization problem by maximizing the ratio of between-class dispersion and within-class dispersion matrix. Then, we solve this TL regularized L1-norm optimization problem by using an iterative algorithm to generate a projection matrix. The projected reduced dimension HSI data are further used for classification by using SVM classifier. %We analyze the classification performance by applying it over the spectral information of two real world HSI datasets.

% The rest of the paper is organized as follows. In section~\ref{sec:prelimnary}, a brief introduction to the conventional L2-norm based SC method is discussed. We present the proposed L1-SC method including its objective function and algorithmic procedure for its solution in section~\ref{sec:proposed_work}. Then section~\ref{sec:res_analysis} enumerates the experimental results of the proposed L1-SC method over two HSI datasets. Finally, we give the conclusive remarks to our work in section~\ref{sec:conclusn}.   

\section{Related Works} \label{sec:prelimnary}
 
  The classical LDA method is computed using the class centers of the data, which means it assumes that, the distribution of the data is Gaussian with equal variance. However, the characteristics of real-world HSI data is multimodal, heteroscedastic with more complex distribution than Gaussian \cite{zhang2013semisupervised}. %That brings the failure of LDA on these multimodal HSI data. 
The scaling cut (SC) method is proposed in \cite{zhang2015scaling,zhang2013semisupervised} to overcome this limitation of LDA by constructing the dissimilarity matrix among the data samples.    

  Let $X =\{x_i,L_i\}|_{i=1}^n \in R^{D\times n}$ is the input training dataset. Here $L_i = \{1,2,...,C\}$ is the class label of the corresponding training data with total $C$ classes and $n$ training data samples. The objective of DR methods is to determine a projection matrix $W$ that project the input training data $X$ of dimensions $D$ into reduced dimension $d$, i.e., $Y = W^T X$ such that $Y \in R^{d \times n}$, $W \in R ^{D \times d}$, and $d << D$. The between-class dissimilarity matrix and the within-class dissimilarity matrix of SC are defined as
\begin{equation}
\begin{split}
S_{B_k}\, = \,\sum\limits_{x_i\, \in \,{U_k}} {\sum\limits_{x_j\, \in \,{{\bar U}_k}} {\frac{1}{{{n_k}{n_{\bar{k}}}}}\,({x_i} - {x_j}){{({x_i} - {x_j})}^T}} }\\
S_{W_k}\, = \,\sum\limits_{x_i\, \in \,{U_k}} {\sum\limits_{x_j\, \in \,{{U}_k}} {{\frac{1}{{{n_k}{n_k}}}}({x_i} - {x_j}){{({x_i} - {x_j})}^T}} }  
\end{split}
\end{equation}
where $U_k$ represents all the samples from $k$th class and $n_k$ is the total number of elements in $U_k$. Similarly, $\bar{U_k}$ represents all the data points that does not belong to the $k$th class and ${n_{\bar{k}}}$ denotes the total number of elements in $\bar{U_k}$. $S_{B_k}$ represents the dissimilarity matrix between $U_k$ and $\bar{U_k}$, whereas $S_{W_k}$ is the dissimilarity matrix within the $U_k$ class. Based on the $S_{W_k}$ and $S_{B_k}$, the objective function $J(W)$ of SC can be written as
\begin{align}\label{eq: scaling_cut}
\begin{aligned}
J(W) \, & = \,\frac{{\left| {\sum\limits_{k\, = \,1}^c {{W^T}{S_{B_k}}W} } \right|}}{{\left| {\sum\limits_{k\, = \,1}^c {({W^T}{S_{W_k}}W\, + {W^T}{S_{B_k}}W\,)} } \right|}}\\
 & = \,\frac{{\left| {{W^T}{S_{B}}W} \right|}}{{\left| {{W^T}{({S_{W}} + {S_{B}})W}} \right|}}
 = \,\frac{{\left| {{W^T}{S_{B}}W} \right|}}{{\left| {{W^T}{S_{T}}\,W} \right|}}
\end{aligned}
\end{align}
where, $S_{B}=\sum\nolimits_{k=1}^C{S_{B_k}}$ is denoted as the between-class, $S_{W}=\sum\nolimits_{k=1}^C{S_{W_k}}$ the within-class, $S_{T} = (S_{B} + S_{W})$ is the total-class dissimilarity matrix and $W$ is the projection matrix. %The dissimilarity matrix is scaled according to the size of the class. Hence, this graph cut is termed as scaling cut.

% 	\subsection{Conventional L2- Norm based Graph Local Scaling Cut Criterion Revisited}
    
%     Suppose there are $N$ training samples with $D$ dimensions $x_i \in R^D$, $i = 1,2, ... , N$ with labels $L_i|_{i=1}^N \in \{1,2,...,C \}$ and $C$ is the number of classes. The input training data set is denoted by $X = \{x_i, L_i\}|_{i=1}^N \in R^{D \times N}$. In order to reveal the intrinsic geometry and manifold structure of the data, we construct a between-class dissimilarity matrix $S_b^{spec}$ and within-class dissimilarity matrix $S_w^{spec}$, given by %considering the spectral-domain information. The     
% \begin{align}
% S_b^{spec} = \sum _{c=1}^C {\sum _{i=1}^{N_c}} \sum _{j=1}^{N} N_{ij}^b(x_i-x_j)(x_i-x_j)^T \\
% S_w^{spec} = \sum _{c=1}^C {\sum _{i=1}^{N_c}} \sum _{j=1}^{N} N_{ij}^w(x_i-x_j)(x_i-x_j)^T
% \end{align}
% where $N^b(\cdot)$ represents the between-class $k_b$ nearest neighborhood and $N^w(\cdot)$ is within-class $k_w$ nearest neighborhood. Here, $N_c$ is the number of elements in the $c^{th}$ class, $N_{ij}^b = {1}/{N_c k_b}$ only if $x_j \in N^b(x_i)$ otherwise $N_{ij}^b=0$ and $N_{ij}^w = {1}/{N_c k_w}$ only if $x_j \in N^w(x_i)$ otherwise $N_{ij}^w=0$.
  
 % 	\subsection{Outline of L1-norm SC}
  
  \section{Proposed Trace Lasso Regularized L1-GC}
  
  The proposed method initially use the guided filter \cite{kang2014spectral} to incorporate the spatial correlation with the spectral correlation by maintaining the pixel consistency. The discriminant information in the filtered data is extracted by the proposed TL-L1GC. 
  %followed by apply the proposed DR algorithm and classifier.    
  %The spectral bands of the HSI data are highly correlated and redundant. Furthermore, 
%  The high spatial variability in HSI data produce an undesired classification performance. To overcome this, a guided filter \cite{kang2014spectral} is used here to spatially maintain the pixel consistency. Initially, the guided filter is applied on the HSI data followed by the proposed DR algorithm.    
  
  %\subsection{Problem formulation}
  
%  In this subsection, we will present our proposed trace lasso regularized L1-norm based graph cut (TL-L1GC) method. 
  The conventional L2-norm based graph cut methods mostly concentrate on the geometric structure and correlation between the data points. However, they are prone to noises and outliers. Although L1-norm is a robust alternative to it, they omit the data correlation, that lead to unstable results. To mitigate the above all issues, we focus on robust trace lasso regularized L1-norm based graph based scaling cut criterion (L1GC), which adaptively balance the noise sensitivity along with the correlation among the data.  
    %\textbf{The conventional L2-norm based graph cut criterion enhances the between-class compactness and within-class dissimilarity among the data points to obtain the projection matrix. These L2-norm based models emphasize the geometric structure and correlation between the data points. The square euclidean distance in L2-norm exaggerate the effect of sensitivity to outliers and noises \cite{wang2014fisher}, \cite{liu2017non}. To address this issue, L1-norm is considered as the robust alternative. Although, L1-norm based models are robust to outliers and noises, they overlook the correlation among the data points. That lead to unstable results. In order to mitigate the above all issues, we focus on robust trace lasso regularized L1-norm based graph based scaling cut criterion (L1GC). [repeated in introduction]} 
    This formulation replaces the L2-norm optimization by L1-norm optimization. Then to balance the L1 and L2 norm and considering sparsity and correlation simultaneously, it utilizes the trace lasso to regularize the basis vector. This trace lasso regularized L1-norm based SC method is solved by following an iterative algorithm     
    
%    Some researchers utilize the L1-norm instead of L2-norm to develop a robust algorithms \cite{wang2014fisher}, \cite{zhong2014discriminant}.
    
 %   These outlier elements drift the projection vectors from the desired projection directions. Hence, it reduces the flexibility of L2-norm based algorithms. To handle this issue, L1-norm based technique is widely used as a robust alternative of conventional L2-norm based technique. Motivated by the idea of L1-norm based modeling, we propose to model the graph based scaling cut criterion by using the L1-norm optimization instead of L2-norm optimization. This L1-norm based SC method is solved by following iterative algorithm
    
    Inspired by the existing literature on L1-norm based scaling cut method in \cite{mohanty2017graph} and trace lasso regularization in \cite{grave2011trace} and \cite{lu2016l1}, we propose to maximize the SC criterion using L1-norm with trace lasso regularizer. The equation (\ref{eq: scaling_cut}) can be simplified to a trace difference \cite{fukunaga2013introduction}, \cite{zhang2015scaling} problem, which can further be reduced to the Frobenius norm based trace difference problem, given by,
     
\begin{align}
\begin{aligned}
&W^* = \mbox{arg} \max_{W^TW = I} \frac{Tr(W^TS_{B}W)}{Tr(W^TS_{W}W)} \\
&= \mbox{arg} \max_{W^TW = I} {Tr(W^T(S_{B} - \lambda S_{W})W)}  \\
&=\mbox{arg} \max_{W^TW = I} {(Tr(W^TS_{B}W) - Tr(W^TS_{W}W))}  \\
%&=\mbox{arg} \max_{W^TW = I} {||(W^TS_{B})||_F^2 - ||(W^TS_{W})||_F^2} \\
&= \max_{W^TW = I}{\sum\limits_{\substack{k; x_i \in U_k;\\x_j \in \bar{U}_k}}\frac{1}{n_k n_{\bar{k}}} Tr\Big(W^T({x_i} - {x_j}){{({x_i} - {x_j})}^T}W\Big)}  \\
&\qquad-{\sum\limits_{\substack{k; x_i \in U_k;\\x_j \in U_k}} \frac{1}{n_k n_{k}}  Tr\Big(W^T({x_i} - {x_j}){{({x_i} - {x_j})}^T}W\Big)} \\ 
% \end{aligned}
% \end{align}
% \begin{align}
% \begin{aligned}
&= \max_{W^TW = I}{\sum\limits_{k = 1}^c\sum\limits_{x_i \in U_k}\sum\limits_{x_j \in \bar{U}_k} \frac{1}{n_k n_{\bar{k}}} \left\|W^T({x_i} - {x_j})\right\|_F^2} \\ 
&\qquad-{\sum\limits_{k = 1}^c\sum\limits_{x_i \in U_k}\sum\limits_{x_j \in U_k} \frac{1}{n_k n_{k}} \left\|W^T({x_i} - {x_j})\right\|_F^2}  \\ %\nonumber
\end{aligned}
\end{align}

where, $\lambda$ is the eigenvalue parameter. As can be observed, the above objective function is the trace difference problem based on Frobenius norm. As it involves the square operations, it is sensitive to noise and outliers. To reduce the sensitivity and to maintain the correlation, we reformulate the objective function in terms of the L1-norm with trace lasso regularizer. The proposed model of the objective function for L1-norm SC with trace lasso regularizer is defined as,

\begin{align} \label{eq:TL_l1sc_optmiz_prblm}
\begin{aligned}
v_{opt} =& \max_{v^Tv = 1}{\sum\limits_{k\, = \,1}^c\sum\limits_{x_i \in U_k}\sum\limits_{x_j \in \bar{U}_k} \left\|{{{v^T}\,{\frac{1}{{{n_k}{n_{\bar{k}}}}}\,({x_i} - {x_j})}} } \right\|_1} \\
-&{\sum\limits_{k\, = \,1}^c \sum\limits_{x_i , x_j \in {U}_k} \left\| {{{{v^T}\,{\frac{1}{{{n_k}{n_{k}}}}\,({x_i} - {x_j})}}} } \right\|_1}-\delta ||X^T Diag(v)||_* \\
=& \max_{v^Tv = 1}{\sum\limits_{k = 1}^c\sum\limits_{x_i \in U_k}\sum\limits_{x_j \in \bar{U}_k} \frac{1}{{{n_k}{n_{\bar{k}}}}} |{{{v^T}{({x_i} - {x_j})}} }|}\\
-&{\sum\limits_{k = 1}^c \sum\limits_{x_i, x_j \in {U}_k} \frac{1}{{{n_k}{n_{k}}}} |{{{{v^T}{({x_i} - {x_j})}}} }|} - \delta ||X^T Diag(v)||_* \\
=&\mbox{arg} \max_{v^Tv = 1} {||(v^TF)||_1 - ||(v^TG)||_1} - \delta ||X^T Diag(v)||_* \\
\end{aligned}
\end{align} 

and, 
\begin{align} \label{eq:F_G_dispersion}
\begin{aligned}
F = \sum\limits_{k = 1}^c\sum\limits_{x_i \in U_k}\sum\limits_{x_j \in \bar{U}_k} \frac{1}{{{n_k}{n_{\bar{k}}}}} |{{{v^T}{({x_i} - {x_j})}} }| \\
G = \sum\limits_{k = 1}^c \sum\limits_{x_i \in U_k}\sum\limits_{x_j \in {U}_k} \frac{1}{{{n_k}{n_{k}}}} |{{{{v^T}{({x_i} - {x_j})}}} }|
\end{aligned}
\end{align}
where, $F = [f_1, f_2, ...,f_n] \in R^{D \times n}$ is the between-class dispersion, $G = [g_1, g_2, ..., g_n] \in R^{D \times n}$ is the within-class dispersion, $||X^T Diag(v)||_*$ is the trace lasso, $\delta$ is the regularizing parameter, $||\bullet||_1$ denote the L1-norm and $||\bullet||_*$ denotes the trace norm of the matrix (trace lasso), i.e. the sum of its singular values. $Diag(\bullet)$ represents the conversion of a vector to a diagonal matrix. As oppose to $Diag(\bullet)$, we will use $diag(\bullet)$ to represent the diagonals of a matrix. The term $X$ in trace lasso $(||X^T Diag(v)||_*)$ is the sample data matrix. As per \cite{grave2011trace}, the trace lasso interpolate between L2 and L1 norm to consider both sparsity as well as correlation of the data simultaneously. The optimized projection vector ($v_{opt}$) in (\ref{eq:TL_l1sc_optmiz_prblm}) is used to construct the optimal projection matrix $ V = \{v_1, v_2, ..., v_d\}$. These projecting vectors are sequentially optimized in $d$ directions. The following iterative algorithm is derived to find the optimal projection vector $v$. The entire optimization procedure for TL-L1GC method is listed below.

\subsection*{Optimization procedure of TL-L1GC}
%In this subsection, we will discuss the procedure to find the optimal vector $v_{opt}$. 
We used an iterative algorithm to determine the local optimal $v_{opt}$ by maximizing the equation (\ref{eq:TL_l1sc_optmiz_prblm}). The basic idea of the proposed technique is to iteratively update the $v_{opt}$ until it converges to its local optima.

As per \cite{grave2011trace} and  \cite{lu2016l1}, the trace norm of a matrix $J \in R^{n \times d}$ is computed as 
\begin{equation} \label{eq:trace_norm}
||J||_* = \frac{1}{2}\, \begin{array}{*{20}{c}}
{\inf }\\
{s \succ \underbar{0}}
\end{array} tr(J^T S^{-1} J) + tr(S) 
\end{equation}
 where, $tr(\bullet)$ represent the trace of the matrix and the infimum of the matrix is achieved for $S = (J^T J)^{1/2}$.  

By using the equation~(\ref{eq:trace_norm}), we can reformulate the objective function in equation~(\ref{eq:TL_l1sc_optmiz_prblm}) as %\textbf{(why F is replaced by $S_B$)}
\begin{align}
\begin{aligned} \label{eq:reform_obj_func}
\mbox{arg} \max_{v,S}\, & {||(v^TF)||_1 - ||(v^TG)||_1} \\
& - \frac{\delta}{2} v^T Diag(diag(XS^{-1}X^T))v - \frac{\delta}{2} tr(S)
\end{aligned}
\end{align}
The maximum value of the equation (\ref{eq:reform_obj_func}) is achieved for $S = (X^T Diag(v)^2 X)^{1/2}$. The variable $S$ and $v$ in the objective function are optimized alternatively: updating $S$ with $v$ fixed, and updating $v$ with $S$ fixed. Here $v$ is updated with fixed $S$ in the following equation %update the $v$ by fixing the $S$ % Initially $S$ is computed by considering the initialized $v_0$, later $S$ is fixed to determine the optimal $v$ by solving the following equation 
\begin{align} \label{eq:solve_for_v}
\begin{aligned}
\mbox{arg} \max_{v,S}\, &{||(v^TF)||_1 - ||(v^TG)||_1} \\ 
&- \frac{\delta}{2} v^T Diag(diag(XS^{-1}X^T))v
\end{aligned}
\end{align}
 The objective function in (\ref{eq:solve_for_v}) is similar to the trace difference formulation of the generalized graph cut in \cite{zhang2015scaling}. The regularization term in the objective function makes it difficult to determine the optimal $v$. Inspired by the current literature \cite{mohanty2017graph} and \cite{lu2016l1}, we are using trace lasso regularized L1-norm optimization technique in this work. Thus, we solve the objective function (\ref{eq:solve_for_v}) to find the optimal projection vector $v^*$ by the iterative technique. The algorithmic procedure of TL-L1GC is given as follows.      

\begin{itemize}
\item[1.]  The iteration variable $t$ is set to zero ($t=0$). Then we randomly initialize the $d$ dimensional vector $v(t)$ and normalize it such that ${v(t)}^T{v(t)} = 1$.
\item[2.]  Initially, we compute the inverse of $S$, i.e. $S^{-1}$, by using the following equation
\begin{align} \label{eq:s_inv}
S^{-1} = U Diag(1/\sqrt{s_i})V
\end{align}
where, $U$ and $V$ are the left and right eigen vectors and $s_i$ is the $i$th ($i=1,2,...,n$) eigen value of $X^TDiag(v(0))^2X$. Here, $U$ and $V$ are orthogonal matrices. The computed $S^{-1}$ is fixed and used to update the $v(t)$ to $v(t+1)$ in the following steps. 
\item[3.] Two sign functions are defined to compensate the absolute value operation for between-class and within-class dispersion term of (\ref{eq:solve_for_v}) . These sign functions are computed as
\begin{equation}\label{eq:sign_func}
\begin{split}
{q_{i}}(t) = & \,\left\{ {\begin{array}{*{20}{c}}
{1,\,\,\,\,\,\,\mbox{if}\,\,\,\,{v^T}(t)f_i\,\, > 0}\\
{ - 1,\,\,\,\,\,\mbox{if}\,\,\,{v^T}(t)f_i\, \le 0}
\end{array}} \right. i = 1,2,....,n \\
 & \mbox{and} \\
 {r_{j}}(t) = & \,\left\{ {\begin{array}{*{20}{c}}
{1,\,\,\,\,\,\,\mbox{if}\,\,\,\,{v^T}(t)g_j\,\, > 0}\\
{ - 1,\,\,\,\,\,\mbox{if}\,\,\,{v^T}(t)g_j\, \le 0}
\end{array}} \right.
j = 1,2,....,n
\end{split}
\end{equation}
\item[4.] Then the $v(t)$ is updated to $v(t+1)$ based on the Theorem~1 in \cite{lu2016l1}. It is given as 
\begin{equation} \label{eq:update_v}
v(t+1) = (\delta Diag(diag(X^TS^{-1}X) + M(t)) )^{-1} N(t)
\end{equation}
where $\delta$ is the regularizing parameter, $N(t) = \sum\limits_{i = 1}^n {{q_{i}}(t)f_i}$, and $M(t) = \sum\limits_{i=1}^n {\frac{{g_i}g_i^T}{|z_i(t)|}}$, here $z_i(t) = v(t)^Tg_i$.  
\item[5.] Convergence check: If the objective function at $v(t)$ doesn't show significant increment on $v(t+1)$ or $||J\{v(t+1)\}-J\{v(t)\}|| \le \epsilon$ or total iteration number is greater then maximum given iteration number, then go to step-7, otherwise go to step-3. %\textbf{(should it be reversed. goto step 7, otherwise goto step 3???)}
\item[6.] Stop iteration and assign $v^* = v(t+1)$ and update train data by
\begin{equation} \label{eq:update_traindata}
X \leftarrow X - {v^*}({v^*}^T)X
\end{equation}
\item[7.] Append the projection vector $v^*$ in the column space of the projection matrix $V = [V,v*]$.
\item[8.] If the iteration $< d$, then goto step~2. Here the updated train data and $v(t+1)$ are used to update the $S^{-1}$ in step 2. Further, the computation of $v(t+2)$ is depended upon the updated $S^{-1}$. Thus, the algorithm will end up with $V$ such that $V \in R^{D\times d}$.
\end{itemize}

Using the above procedure, we can form the optimal projection matrix $V$ of size $R^{D \times d}$. The pseudo-code for the complete algorithmic procedure for the projection matrix of TL-L1GC is listed in Algorithm~$1$

\begin{algorithm}[H]
\caption{Trace lasso regularized L1GC algorithm}\label{algo}
\begin{algorithmic}[1]

\Require {The training dataset $ \{x_i, L_i\} _{i = 1}^n \in R^{D \times n}$; $L_i$ is the label of each training data $x_i$; Desired dimensionality is $d$ and $d \ll D$.}
\State Formulate the trace lasso regularized L1-norm based graph cut objective function in (\ref{eq:TL_l1sc_optmiz_prblm}) to solve the optimization problem. %$\mathit{TrustedData} \gets \emptyset$
\State Compute $F=[f_1, f_2, ..., f_n]$ and $G=[g_1, g_2, ..., g_n]$ using (\ref{eq:F_G_dispersion}).
\State Set $t = 0$, $iter = 0$, projection matrix $V=[\,]$ and initialize $v(0)$ such that ${v(0)}^Tv(0)=1$
\ForAll{\{1...d\}}
    \State Compute the $S^{-1}$ using Eq.~(\ref{eq:s_inv})
    \While {true}
    \State $iter = iter + 1$
    \State Compute the sign function $q_i(t)$ and $r_j(t)$ using Eq.~(\ref{eq:sign_func})
    \State Compute $Fv = ||v(t)^T F||_1$ and $Gv = ||v(t)^T G||_1$
    \State Update the $v(t+1)$ by using Eq.~(\ref{eq:update_v})
    \State Use $v(t+1)$ to determine $Fv' = ||v(t+1)^T F||_1$ and $Gv' = ||v(t+1)^T G||_1$
    \State Converge if: $||v(t+1)-v(t)|| \le \epsilon$ or $iter > itmax$
    \State $v(t) \leftarrow v(t+1)$
    
%       \ForAll{$c\in\mathit{TrustedCode}$}
%         \State $d\gets d\cup\sfunction{CFDataUsed}(c)$
%         \State $d\gets d\setminus \mathit{TrustedCode}$
%       \EndFor
%       \ForAll{$\mathit{ptr}\in d$}
%         \If{$\sfunction{CodeIsValid}(\text{code at }\mathit{ptr})$}
%           \State add code at $\mathit{ptr}$ to $\mathit{TrustedCode}$
%           \State add $\mathit{ptr}$ to $\mathit{TrustedData}$
%         \Else
%           \State{raise alarm}
%         \EndIf
%       \EndFor
%      \State $\sfunction{MonitorForWrites}(\mathit{TrustedCode}\cup\mathit{TrustedData})$
    \EndWhile
\State \textbf{endwhile}
\State Update the input data $X$ by using Eq.~(\ref{eq:update_traindata})
\State Pad these optimal projection vectors $v^*$ into the optimal matrix by $V=[V,{v^*}]$ $\in R^{D \times d}$
\State \textbf{endfor}
\EndFor
\Ensure {Projection matrix $V=\{v_1, v_2, ..., v_d\}$ $\in R^{D \times d}$, consists of $d$ projection vectors}
\end{algorithmic}
\end{algorithm}

\section{Experiments}
% 	\subsection{Databse Description}
    This section evaluates the performance of the proposed TL-L1GC method on two benchmark HSI datasets\footnote{http://www.ehu.eus/ccwintco/index.php?title=Hyperspectral\_Remote\\\_Sensing\_Scenes}: Botswana $(D=145,  C=14)$ and Salinas $(D=204, C=16)$. Here, we have made a comparative study with the state-of-the-art L2 based methods (such as SC \cite{zhang2015scaling}, LSC \cite{zhang2013semisupervised}) and L1 based methods (e.g. L1-LDA \cite{wang2014fisher}, L1-SC \cite{mohanty2017graph}). The conventional L2-norm based methods use PCA as preprocessing step. In these experiments, $10$ samples are randomly selected from each class of the dataset for training and the remaining data samples are used for testing. All the results are obtained by taking the average of the $5$ iterations. In order to identify the robustness of the proposed algorithm, SVM with a linear kernel is used as the classifier. Empirically, the regularizing parameter $\delta$ is tuned to $0.4$ at varied dimensions ranges from $4$ to $60$ with data samples $10$.
    %The regularizing parameter $\delta$ is tuned from the set $\{0.1,0.2,...,0.9\}$. After experimenting over the aforementioned range of values for dimensions $10$ and data samples $10$, we found that the maximum overall classification accuracy (OA) is achieved at parameter $( \delta = 0.4)$. Then, we use that value of $\delta$ in rest of our experiments for both the datasets.
       
The proposed TL-L1GC method is compared with other state-of-the-art L2-norm and L1-norm based methods. The statistics of the highest OA along with corresponding average accuracy (AA), kappa coefficient ($k$) and dimension of the algorithms for Botswana and Salinas dataset are highlighted in Table~\ref{tab:Classfcn_L1_l2}. Here the classification results are obtained by performing the average of $5$ iterations while using $10$ random training samples per class. 
%In order to show the robustness of the algorithm, we have also considered the class average accuracy and kappa coefficients along with the overall classification accuracy as the performance measure. Table~\ref{tab:Classfcn_L1_l2} gives the following observations
%\begin{itemize}
%\item The OA of L1-norm based methods performs better than the other state-of-art L2-norm based methods.
%\item 
The classification results of the proposed TL-L1GC method significantly outperforms the other L2-norm as well as L1-norm based methods for both the datasets in all the metrics.
%\item 
In Botswana dataset, the proposed TL-L1GC method outperforms the other methods by a large margin of approximately $10\%$ of OA at the reduced dimension of $58$. Similarly, for Salinas data set, it also outperforms others at $54$ dimensions. This verifies the effectiveness of the algorithm in finding proper projection direction. 
%\end{itemize}    
    
    \begin{table*}[!htp]
\centering
\caption{Classification performance of proposed approach compared with other L2-norm and L1-norm based approaches }
\label{tab:Classfcn_L1_l2}
\begin{tabular}{l|ccc|c|ccc|c}
\hline
\multicolumn{1}{c|}{Dataset}                & \multicolumn{4}{c|}{\textbf{Botswana}}                                  & \multicolumn{4}{c}{\textbf{Salinas}}                                   \\ \hline 
Methods                                     & OA             & AA             & k              & dim         & OA             & AA             & k              & dim         \\ \hline
SC    \cite{zhang2015scaling}             & $76.81$          & $77.09$          & $59.15$          & $56$          & $81.11$          & $87.57$          & $62.75$          & $58$          \\
LSC    \cite{zhang2013semisupervised}     & $78.12$          & $79.25$          & $60.20$          & $52$          & $82.98$          & $88.84$          & $63.12$          & $50$          \\ \hline
L1-LDA   \cite{wang2014fisher}            & $86.90$          & $88.18$          & $69.02$          & $56$          & $83.66$          & $88.15$          & $61.92$          & $56$          \\
L1-SC   \cite{mohanty2017graph}           & $84.06$          & $85.53$          & $65.95$          & $42$          & $83.57$          & $86.81$          & $61.70$          & $56$          \\ \hline
\textbf{TL-L1GC}    \textbf{(Without Filter)} & $\mathbf{84.92}$ & $\mathbf{85.94}$ & $\mathbf{77.54}$ & $\mathbf{58}$ & $\mathbf{84.54}$ & $\mathbf{87.25}$ & $\mathbf{70.45}$ & $\mathbf{54}$ \\ %\hline
\textbf{TL-L1GC}    \textbf{(With Filter)} & $\mathbf{94.54}$ & $\mathbf{94.69}$ & $\mathbf{75.54}$ & $\mathbf{58}$ & $\mathbf{86.42}$ & $\mathbf{91.67}$ & $\mathbf{67.53}$ & $\mathbf{54}$ \\ \hline
\end{tabular}
\end{table*}
 
 %##################################################################################################################

\begin{figure}[htp]
	\centering
	\begin{subfigure}{.24\textwidth}
		\centering
		\includegraphics[width=.99\linewidth]{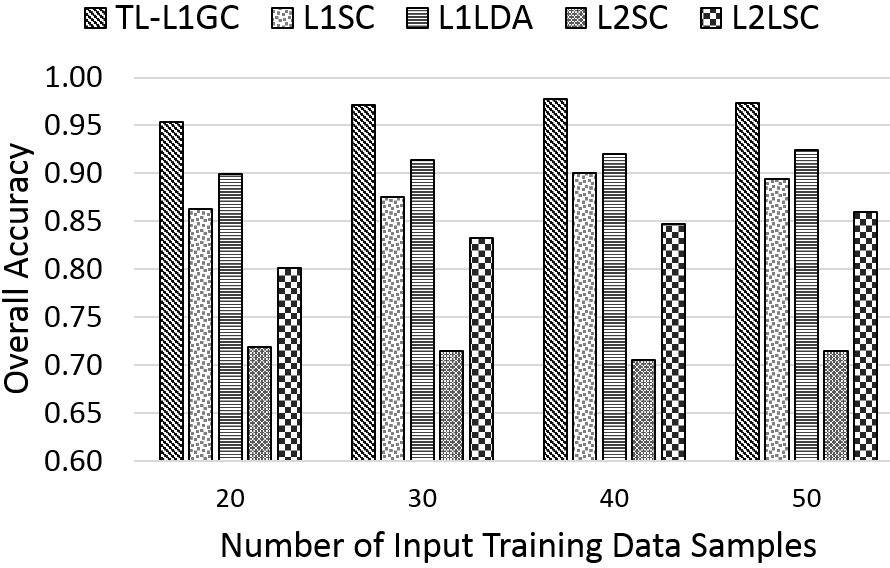}
		\caption{}
		\label{fig:MultiTrnSmpl_Bots}
	\end{subfigure}    
% \end{figure}
% \begin{figure}[h]
  	\begin{subfigure}{.24\textwidth}
		\centering
		\includegraphics[width=.99\linewidth]{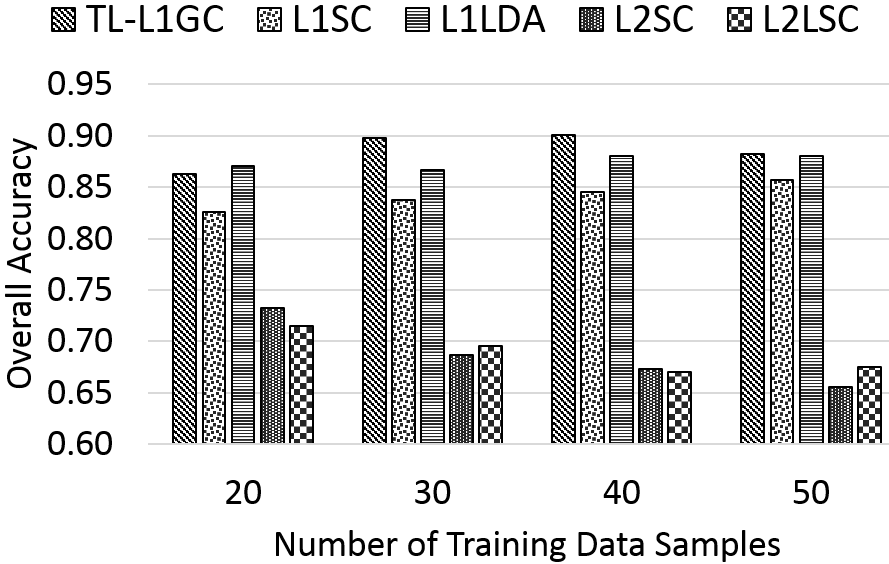}
		\caption{}
		\label{fig:MultiTrnSmpl_salina}
	\end{subfigure}
	\caption{ Illustration of performance of the proposed method with respect to other methods in terms of overall classification accuracies for different input training samples ($20, 30, 40$ and $50$) on two HSI datasets Botswana (\ref{fig:MultiTrnSmpl_Bots}) and Salinas (\ref{fig:MultiTrnSmpl_salina}).}
	\label{fig:Bots_salina_TrnSmpl}
\end{figure}
%##############################################################################################################
 
 Fig.~\ref{fig:Bots_salina_TrnSmpl} illustrates the behavior of different L2-norm and L1-norm based algorithms in terms of overall classification accuracy with respect to varied number of input data samples. From this figure, it is clearly observed that the proposed TL-L1GC method significantly outperforms the other L1-norm and L2-norm based methods when the input training data sample size is less.
 
\begin{figure}[htp]
	\centering
	\begin{subfigure}{.24\textwidth}
		\centering
		\includegraphics[width=.99\linewidth]{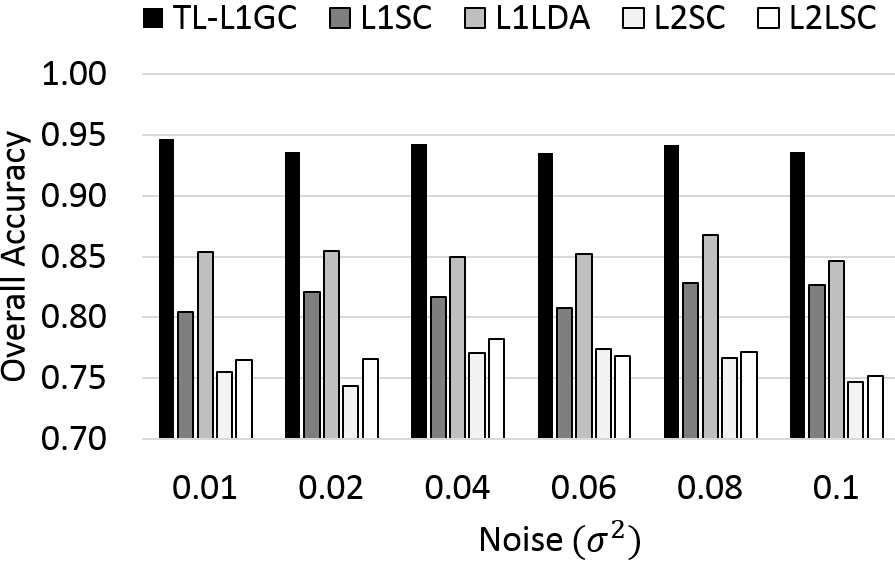}
		\caption{}
		\label{fig:noise_Bots}
	\end{subfigure}    
% \end{figure}
% \begin{figure}[h]
  	\begin{subfigure}{.24\textwidth}
		\centering
		\includegraphics[width=.99\linewidth]{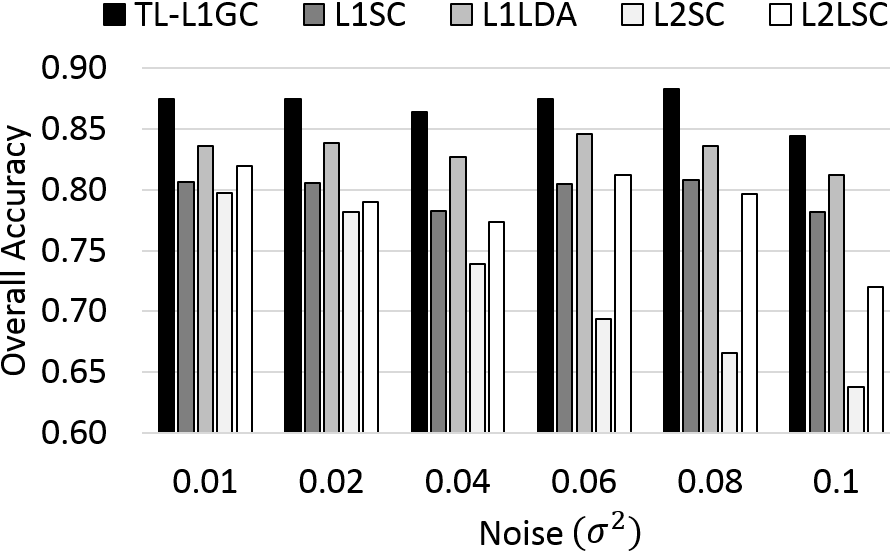}
		\caption{}
		\label{fig:noise_Salina}
	\end{subfigure}
	\caption{ Illustration of noise robustness of the proposed method with respect to other methods on two HSI datasets Botswana (\ref{fig:noise_Bots}) and Salinas (\ref{fig:noise_Salina}).}
	\label{fig:Bots_salina_noise}
\end{figure}

 In order to, demonstrate the noise robustness features of the proposed TL-L1GC method with respect to others, we inject white Gaussian noises of different levels to the raw input HSI data and performed the classification on these data using different approaches. The variance of the noise level is varied from 1\% to 10\% of the variance of the pixel values at fixed dimension $50$ and number of training sample $10$. As observed in Fig.~\ref{fig:Bots_salina_noise}, the proposed TL-L1GC method is more robust to noises and achieve better classification accuracies than other methods.

 %##################################################################################################################

\section{Conclusion}
In this study, we have proposed a novel DR method TL-L1GC preceded by a guided filter through computing the trace lasso regularized L1-norm based graph cut. This method determines the projection directions by adaptively considering both the sparsity as well as data correlation in both spectral and spatial domain. It exploits the discriminant structure while preserving the geometrical structure of the data. Our method differs from other state-of-the-art methods in various ways. For instance, it maintains the intrinsic property as well as the distribution of the data, and we believe it handles multimodal and heteroscedastic noisy data with high correlation quite well. We examined the performance of our method and other methods over two real-world HSI datasets. The promising results of TL-L1GC on these two datasets demonstrates its noise robustness and efficiency.

\bibliographystyle{IEEEtran}
% argument is your BibTeX string definitions and bibliography database(s)
\bibliography{mybib}

% \begin{thebibliography}{1}

% \bibitem{IEEEhowto:kopka}
% H.~Kopka and P.~W. Daly, \emph{A Guide to \LaTeX}, 3rd~ed.\hskip 1em plus
%   0.5em minus 0.4em\relax Harlow, England: Addison-Wesley, 1999.

% \end{thebibliography}

% \begin{IEEEbiography}{Michael Shell}
% Biography text here.
% \end{IEEEbiography}

% % if you will not have a photo at all:
% \begin{IEEEbiographynophoto}{John Doe}
% Biography text here.
% \end{IEEEbiographynophoto}

% % insert where needed to balance the two columns on the last page with
% % biographies
% %\newpage

% \begin{IEEEbiographynophoto}{Jane Doe}
% Biography text here.
% \end{IEEEbiographynophoto}

\end{document}